\documentclass[a4paper,10pt,conference]{ieeeconf}
\usepackage{FG2027}
\FGfinalcopy
\IEEEoverridecommandlockouts
\usepackage{amsmath,amssymb,graphicx,booktabs,xcolor,placeins,multirow,xspace}
\usepackage{url}
\usepackage[labelfont=bf,font=small]{caption}

\makeatletter\let\NAT@parse\undefined\makeatother
\definecolor{linkblue}{rgb}{0.21,0.49,0.74}
\usepackage[hidelinks,colorlinks,breaklinks,allcolors=linkblue]{hyperref}

\hypersetup{
  pdftitle={Beyond Temporal Overlap: Consumer-Aware Evaluation of Sign-Language Segmentation across Three Corpora},
  pdfauthor={Guohou Zhang and Kalin Stefanov}
}

\newcommand{\mFoneB}{\ensuremath{\mathrm{mF1B}}\xspace}
\newcommand{\mFoneS}{\ensuremath{\mathrm{mF1S}}\xspace}
\newcommand{\WERall}{\ensuremath{\mathrm{WER}_{\mathrm{all}}}\xspace}
\newcommand{\WERiv}{\ensuremath{\mathrm{WER}_{\mathrm{IV}}}\xspace}

\begin{document}
\title{\LARGE \bf Beyond Temporal Overlap: Consumer-Aware Evaluation\\
of Sign-Language Segmentation across Three Corpora}
\author{Guohou Zhang$^{1}$ \qquad Kalin Stefanov$^{2}$\\
{\small $^{1}$Monash University, Australia \qquad
$^{2}$Japan Advanced Institute of Science and Technology, Japan}\\
{\tt\small gzha0067@student.monash.edu \qquad kalins@jaist.ac.jp}}
\maketitle

\begin{abstract}
Sign-language segmentation divides continuous signing into clips. In
this study, the downstream consumer is a frozen recognizer that assigns one
sign label to each predicted clip. Existing metrics instead compare predicted
clips with manually annotated boundaries. They do not directly measure whether
a clip preserves the visual evidence needed by the recognizer or whether the segmenter
produces too many clips. We therefore ask whether overlap metrics and this
consumer prefer the same segmenter. We evaluate 27 outputs
from three segmentation architectures, three random seeds, and three corpora.
Controlled boundary edits separate the effect of removing sign evidence from
that of adding surrounding context. Overlap predicts recognition on average,
but it does not always determine model choice. At the same
intersection-over-union, different recognizers can prefer the shorter or the
longer clip. The real outputs reveal two operating regimes. On DGS, overlap,
recognition, and sequence error select the same model. On BSL,
recognition-based selection recovers and recognizes more reference signs but
increases word error rate (WER). Dicta-Sign shows the reverse trade-off: WER
selection reduces in-vocabulary WER by 31.17 points but misses more recognizable
signs. Segmenters should therefore be evaluated on four axes: temporal
agreement, reference recovery, clip recognition, and sequence error, rather
than on overlap alone.
\end{abstract}

\section{Introduction}
\label{sec:introduction}

Sign-language segmentation divides continuous signing into lexical clips.
These clips support online recognition \cite{zuo2024online}, systems that detect
each sign before classifying it \cite{zhao2026mhb}, lexical mining
\cite{varol2021read}, subtitle alignment \cite{jiang2026sea}, and annotation
\cite{woll2022humans}. In each application, a downstream system receives the
frames inside every predicted interval. We call that system the
\emph{consumer}. A useful segmentation must therefore do more than locate a
sign: it must deliver clips that serve its consumer.

Current benchmarks mainly compare predictions with manual annotations.
Boundary F1 (\mFoneB) checks whether predicted endpoints fall near annotated
endpoints. Segment F1 (\mFoneS) matches intervals at several
intersection-over-union (IoU) thresholds. Frame metrics compare foreground
occupancy \cite{renz2021segmentation,moryossef2023seg,low2025handson}. These
scores measure temporal agreement. They do not directly measure what a
recognizer can recover from the predicted clips.

One example makes the gap concrete. Shrinking an interval removes frames from
the sign. Expanding it adds transition frames or content from a neighboring
sign. Both clips can receive the same IoU, yet a recognizer sees different
evidence. Segment count creates a second problem. Emitting more clips may
recover more signs, but extra clips become insertion errors. Emitting fewer
clips reduces insertions but may miss recognizable signs. An overlap score
cannot describe both effects by itself.

\begin{figure*}[t]
  \centering
  \includegraphics[width=\textwidth]{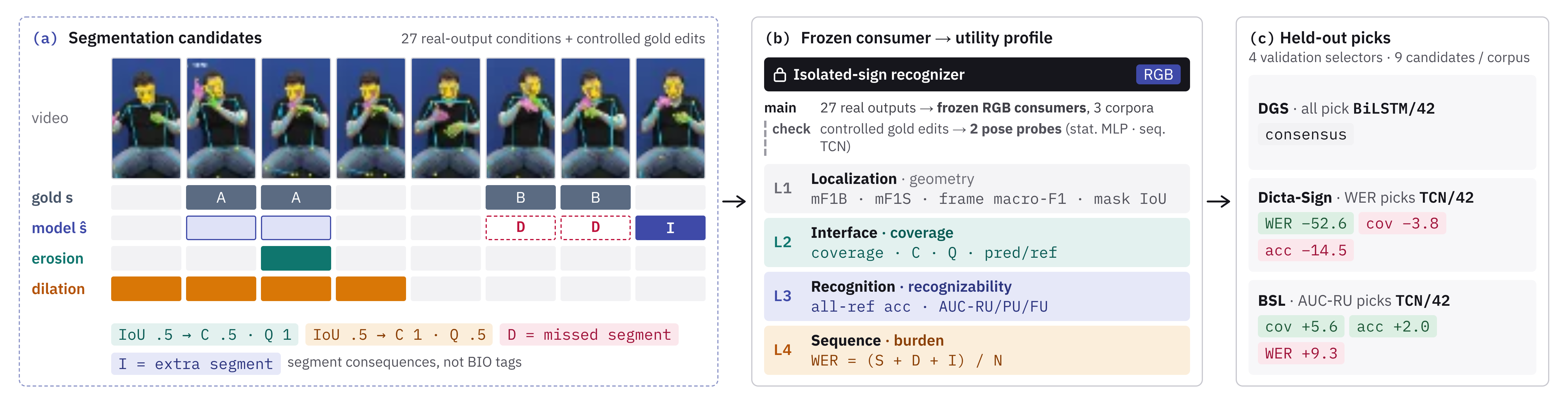}
  \caption{Consumer-aware evaluation of sign segmentation. A Public DGS Corpus
  sequence \cite{hanke2020dgscorpus} connects real model outputs and controlled
  boundary edits to localization, recognition, and sequence-level evaluation.
  The protocol fixes features, vocabularies, matching, and consumers before
  comparing segmenters.}
  \label{fig:overview}
\end{figure*}

We ask a practical question: does the segmenter with the better overlap score
also produce better clips for recognition? We answer it in two ways. First, we
edit gold intervals while controlling either IoU or the number of changed
frames. This experiment shows how removing sign evidence and adding context
affect fixed recognizers. Second, we test whether overlap and recognition select
the same system among real segmenter outputs. Figure~\ref{fig:overview}
summarizes this evaluation.

The study covers three corpora and 27 real outputs: three segmentation
architectures trained with three random seeds on each corpus. Every candidate
uses the same framewise decoding rule. Within a corpus, the vocabulary,
matching rule, and recognition model also remain fixed. We additionally test
the DGS outputs with two pose-based recognizers. Controlled experiments on DGS
and Dicta-Sign apply six common segmentation errors to gold intervals:
erosion, dilation, jitter, deletion, merge, and split.

The experiments reveal two regimes. On DGS, overlap and
recognition prefer the same architecture across three recognizers. On BSL, the
model that recovers and recognizes more reference signs also produces more
sequence errors. Dicta-Sign shows the same trade-off from the other direction:
selecting the model with lower sequence error loses recognizable signs. Removing predictions
linked to out-of-vocabulary signs does not change that selection. These results
show why one overlap score cannot choose a segmenter for every downstream use.

This work makes three contributions:
\begin{itemize}
\item A controlled test showing that equal temporal overlap can yield different
recognition outcomes;
\item A three-corpus benchmark of 27 real segmenter outputs evaluated with
fixed recognizers and validation-only model selection; and
\item An evaluation protocol that reports temporal agreement together with
reference recovery, recognition, and sequence error.
\end{itemize}

The protocol makes model choice explicit. Researchers first state which
downstream system will use the clips. They then select on its validation
objective and report what the chosen segmenter recovers, recognizes, and emits.

\section{Related Work}

\noindent\textbf{Lexical sign segmentation.}
Renz et al. established a reproducible lexical segmentation benchmark using
spatiotemporal features, temporal convolution, and boundary- and segment-level
F1 scores \cite{renz2021segmentation}. Later work expands both the label space
and training regime. Changepoint-modulated pseudo-labels adapt a segmenter to
unlabelled target videos \cite{renz2021changepoint}. Moryossef et al. combine
BIO labels with linguistic
cues to model sign and phrase boundaries on DGS \cite{moryossef2023seg}.
Hands-On couples BIO prediction with Transformer temporal modeling and hand
representations \cite{low2025handson}. Zhao et al. combine skeletal motion and
handshape cues to detect sign boundaries before isolated recognition
\cite{zhao2026mhb}. Human-facing evaluation asks whether an
extracted interval remains intelligible as one sign, complementing temporal
tolerance with functional judgment \cite{woll2022humans}.

BEACON combines BIOE labeling, boundary-knowledge distillation, and structured
decoding, and reports both segmentation scores and downstream WER
\cite{zhang2027beacon}. Its architecture analysis finds that \mFoneS follows WER
more consistently than \mFoneB among the tested models. That result leaves a
broader evaluation question: when do overlap, recognition, and sequence error
select the same system, and what does each objective cost when they disagree?

\noindent\textbf{Continuous and clip-based sign interfaces.}
Modern continuous sign recognition and translation systems encode long frame
sequences directly through RGB and pose streams \cite{chen2022twostream},
visual--language pretraining \cite{zhou2023gfslt}, or contextual
self-supervision \cite{gueuwou2025shubert}. Clip-based systems expose a
different interface. Online CSLR applies an isolated recognizer through a
sliding window \cite{zuo2024online}; lexical mining retrieves sign instances
from co-articulated video \cite{varol2021read}; SEA segments long videos before
aligning clips to subtitles \cite{jiang2026sea}. Annotation systems similarly
depend on intervals that a human can inspect and correct
\cite{woll2022humans}. Subtitle-unit segmentation provides another clip-based
interface at a longer temporal scale \cite{bull2020subtitle}. These consumers
attach different costs to missed signs,
contaminated clips, and additional predictions.

\noindent\textbf{Evaluation of temporal segments.}
Temporal action segmentation already uses several complementary measures. Lea
et al. introduced segmental F1 at IoU thresholds to penalize fragmentation
while tolerating small shifts \cite{lea2017tcn}. Subsequent work reports frame
accuracy, edit score, and F1 across multiple thresholds because each measure
exposes a different error pattern \cite{farha2019mstcn,li2021timestamp}.
Boundary-aware models target over-segmentation hidden by frame accuracy
\cite{ishikawa2021boundary}. OTAS directly compares frame- and boundary-level
measures and pairs automatic scores with a user study \cite{li2024otas}.
Recent time-series evaluation further decomposes segment errors by type
\cite{chavelli2025interpretable}.

Sign-language evaluation faces the same metric-design problem beyond temporal
segmentation. Jiang et al. compare keypoint, embedding, and back-translation
metrics against sign retrieval and human judgments \cite{jiang2025meaningful}.
Their meta-evaluation connects geometric scores to functional evidence; our
protocol applies that principle to the clips produced by a segmenter.
Our protocol asks how many
reference signs are recovered, whether their clips remain recognizable, and
how many errors the complete output sequence creates. Controlled edits explain
why overlap and recognition can disagree. Real model outputs then show whether
the disagreement changes which segmenter is selected.

\section{Consumer-Aware Evaluation}
\label{sec:protocol-v2}

\subsection{Measuring what a segmenter recovers and emits}
We evaluate each segmentation from three views. First, how many reference signs
does it recover? Second, can a fixed recognizer identify the recovered clips?
Third, how many errors does the complete predicted sequence create? The
following measures answer these questions while retaining standard boundary,
segment, and frame scores.

Let $G=\{(s_j,g_j)\}_{j=1}^{N}$ contain the reference intervals and glosses
inside the train-supported consumer vocabulary, and let
$P=\{\hat{s}_i\}_{i=1}^{M}$ contain the emitted intervals. Conventional
evaluation uses the full annotation to compute \mFoneB over 1--4-frame
tolerances, \mFoneS over IoU thresholds $.40:.05:.75$, frame macro-F1, and
foreground IoU. Together, they measure localization.
The \mFoneS grid follows the established segmentation protocol
\cite{renz2021segmentation}. The utility grid is coarser and extends to $.3$ to
measure recognizable clips from permissive through strict overlap; the results
section tests narrower grids and the single $.5$ threshold.

We next match references to predictions. Candidate pairs
with positive temporal overlap are sorted by decreasing IoU, ties are resolved
by prediction/reference indices, and matching stops when no eligible pair
remains. Let $\mathcal{M}\subseteq\{1,\ldots,N\}\times\{1,\ldots,M\}$ denote
the accepted one-to-one matches. \emph{Lexical coverage} is
$|\mathcal{M}|/N$. For each $(j,i)\in\mathcal{M}$,
\begin{equation}
C_{ji}=\frac{|s_j\cap\hat{s}_i|}{|s_j|}, \qquad
Q_{ji}=\frac{|s_j\cap\hat{s}_i|}{|\hat{s}_i|}
\end{equation}
measure retained evidence and temporal purity. $C_{ji}$ decreases when a prediction
removes sign-internal frames; $Q_{ji}$ decreases when it includes surrounding
content. A frozen recognizer assigns one gloss to every predicted clip.
Matched accuracy measures recognition among recovered references.
All-reference accuracy assigns zero to an unmatched reference and therefore
measures recognizable lexical coverage.

A single overlap threshold fixes the strictness of this judgment. We therefore
evaluate multiple thresholds. For threshold $\tau$, let
\begin{equation}
K(\tau)=\sum_{(j,i)\in\mathcal{M}}
\mathbb{1}[\operatorname{IoU}(s_j,\hat{s}_i)\ge\tau]
\mathbb{1}[g_j=\hat g_i].
\end{equation}
Reference utility $R_U(\tau)=K(\tau)/N$ rewards recognizable reference signs.
We integrate this curve by the normalized trapezoidal rule over the predeclared
grid $\tau\in\{.3,.4,.5,.6,.7\}$, producing AUC-RU. Averaging across
partial-to-strict overlap avoids committing model selection to one IoU
threshold. Section~\ref{sec:results-v2} verifies that $.4$--$.6$ AUC and
$R_U(.5)$ preserve every AUC-RU validation selector. The supplement reports
prediction-normalized and harmonic variants as secondary diagnostics.

\begin{table}[t]
\centering
\caption{The selection metric follows the intended use. The remaining measures
show what the selected model gains and loses. S/D/I denotes
substitution/deletion/insertion.}
\label{tab:consumer-map}
\footnotesize
\begin{tabular}{lll}
\toprule
\textbf{Consumer objective} & \textbf{Select with} & \textbf{Report alongside} \\
\midrule
Recover references & AUC-RU & coverage, $C$ \\
Transcribe sequence & \WERall & coverage, S/D/I \\
\bottomrule
\end{tabular}
\end{table}

\subsection{Separating vocabulary limits from segmentation errors}
Sequence scoring aligns the ordered in-vocabulary reference glosses with the
label assigned to every predicted clip. \WERall retains the complete prediction
stream. For a closed-vocabulary recognizer, clips
aligned to an out-of-vocabulary (OOV) sign, unaligned clips, and duplicate
fragments can all become insertions. Values above 100\% identify a stream whose
edit burden exceeds an all-deletion hypothesis.

\WERiv measures the same sequence inside the evaluable vocabulary. At scoring
time, it removes a prediction only when its best positive overlap with an
annotated OOV token exceeds its best overlap with every in-vocabulary token.
Ties and unaligned predictions remain in the hypothesis. Reporting both scores
separates vocabulary scope from segmentation burden: \WERiv evaluates the
train-supported lexical population, while \WERall prices every emitted clip.

The primary consumer assigns a label to every clip. We also test a consumer
that may reject a clip when its maximum softmax probability falls below a
threshold. One corpus-level threshold is fixed from the shared cache of
validation predictions across all nine candidates to retain 95\%, 90\%, or
80\%, then applied unchanged to every test candidate.
A separate unconstrained diagnostic jointly selects the candidate and threshold
that minimize validation \WERiv. The threshold grid contains 0, .01--.05 in
.01 steps, .075, .10, .15, .20, .30--.90 in .10 steps, .95, .975, and .99.
Both choices are then applied once to held-out test data. This diagnostic tests
whether sequence error alone prefers rejecting useful output.

\subsection{Selecting models without test leakage}
Each corpus fixes the vocabulary, consumer, segmenter architectures, and seeds
before test evaluation. The main analysis compares three validation objectives:
\mFoneS, AUC-RU, and \WERall. Prediction-normalized and harmonic utility
selectors are reported in the supplement. Table~\ref{tab:consumer-map} maps the
two downstream objectives to their selection and diagnostic measures. \WERiv
is a vocabulary-scope sensitivity analysis and yields the same selector in all
three corpora. The selected IDs are
written to a manifest and applied once to held-out test. A second comparison
selects the best validation-\mFoneS seed within each architecture, isolating
family effects from seed variation.

Every predeclared selector contrast receives a paired 95\% percentile interval
from 20,000 bootstrap resamples over test signers: 14 on DGS, 6 on Dicta-Sign,
and 21 on BSL. A 19-recording bootstrap accompanies Dicta-Sign. An interval
that excludes zero gives a clear direction. An interval that crosses zero does
not distinguish the two choices at the available cluster count.

\subsection{Creating controlled boundary errors}
Real outputs mix boundary location, segment count, and error type. We isolate
these factors on DGS and Dicta-Sign by editing gold lexical intervals with
erosion, dilation, jitter, deletion, merge, and split. Matched-IoU pairs compare
inner and outer windows at equal
IoU. Equal-frame pairs remove and add the same number of frames. The first
isolates boundary semantics at fixed overlap; the second isolates direction at
fixed edit magnitude. We then count how often \mFoneS and a frozen consumer
prefer opposite conditions.

\section{Experimental Setup}
\label{sec:setup-v2}

\subsection{Three held-out corpus protocols}
Table~\ref{tab:corpus-protocols} summarizes the Public DGS Corpus 3.0.0
uzh-document protocol \cite{hanke2020dgscorpus,zhang2023sltunet}, frontal
single-signer-per-recording views from Dicta-Sign-LSF-v2
\cite{belissen2020dictasign}, and the signer-independent split of Renz et al. for the BSL
Corpus \cite{schembri2013bsl,renz2021segmentation}. DGS removes every document
containing a held-out signer from training. Dicta-Sign retains its canonical
test sessions and draws validation from a disjoint training session; signers
can recur across sessions. BSL follows its official signer-disjoint split.
Frames outside annotated lexical intervals serve as \texttt{O}. Each corpus
retains its native annotation and vocabulary. The BSL recognizer uses the BOBSL
label space \cite{albanie2021bobsl}. We compare models only within the same
corpus.

\begin{table*}[t]
\centering
\caption{Three held-out corpus protocols. Eligible tokens are test references
inside the train-only consumer vocabulary. Absolute recognition values are
compared only within corpus. The final column gives split control and the
number of signer clusters used for test inference.}
\label{tab:corpus-protocols}
\footnotesize
\begin{tabular}{lcccccc}
\toprule
\textbf{Corpus} & \textbf{Train/val/test} & \textbf{Eligible test tokens} & \textbf{Classes} & \textbf{Segmenter input} & \textbf{Frozen consumer} & \textbf{Split / clusters} \\
\midrule
DGS & 583/12/17 videos & 3,080/5,713 (53.9\%) & 500 & MediaPipe 225-D & Logos RGB & signer-disj. / 14 \\
Dicta-Sign & 58/16/20 recordings & 2,018/4,654 (43.4\%) & 183 & MediaPipe 225-D & Logos RGB & session-disj. / 6 \\
BSL & 5,413/763/703 clips & 4,167/4,751 (87.7\%) & 968 & HaMeR 288-D & BOBSL RGB & signer-disj. / 21 \\
\bottomrule
\end{tabular}
\end{table*}

\subsection{Three segmenter families}
The candidate pool spans temporal convolution, bidirectional recurrence, and
shifted-window attention. DGS and Dicta-Sign use 75 MediaPipe Holistic body and
hand landmarks (225 coordinates), centered at the shoulder midpoint and scaled
by shoulder distance. BSL uses the official 288-D HaMeR bilateral hand rotation
features. Missing or non-finite feature blocks are set to zero.

The TCN contains five residual layers with dilations $1,2,4,8,16$, width 192,
and a 63-frame receptive field. The BiLSTM projects each frame to 256 dimensions
and applies four bidirectional layers with 128 units per direction. The
Transformer uses four width-128 layers, eight heads, and alternating 128-frame
windows shifted by 64 frames. Every family predicts the same corpus-specific
BIO labels and is trained with seeds 7, 42, and 2027. Capacity, data splits, and
optimization are fixed across seeds. Training uses class-weighted
cross-entropy, validation macro-F1 checkpoint selection, at most 100 epochs,
and patience 12.

Every checkpoint uses framewise argmax followed by the deterministic local BIO
repair in Hands-On \cite{low2025handson}: an illegal \texttt{I} becomes a valid
local continuation or beginning. Applying the same readout to three families
and three seeds produces nine independently trained candidates per corpus,
without a learned transition or duration model.

\subsection{Frozen recognition consumers}
DGS and Dicta-Sign use corpus-specific RGB recognizers built from a frozen
Logos-pretrained video encoder \cite{ovodov2025logos}. Classification heads and
normalization statistics are fitted on gold training clips. BSL uses a frozen
three-head BOBSL VideoSwin ensemble. The gold-interval Top-1 ceilings are
46.88\%, 65.26\%, and 40.82\% on DGS, Dicta-Sign, and BSL.

DGS and Dicta-Sign use corpus-specific instances of two pose-consumer
architectures. Each instance is trained on gold training clips using its
train-supported vocabulary (500 classes on DGS and 183 on Dicta-Sign), then
frozen before predicted-interval evaluation. Neither consumer is trained or
fine-tuned on predicted intervals or held-out test clips. A statistical MLP
pools mean, standard deviation, extrema, and mean
absolute velocity from each clip; a sequence TCN processes 16 temporally
resampled frames. On DGS, their gold-interval test accuracies are 35.75\% and
39.84\%, respectively. These pose recognizers differ from the RGB recognizer in
both input modality and temporal processing. All three DGS recognizers evaluate
the same nine predicted segment sets. We use the two pose recognizers for the
controlled boundary edits on DGS and Dicta-Sign.

BSL adds an independent three-seed Residual-TCN ensemble over BSL-1K I3D
features. Its 968-class vocabulary matches the primary BOBSL consumer. The
heads are trained on gold training intervals, selected on an inner
signer-disjoint validation split, and frozen before held-out predicted-clip
evaluation; the public test split is not used for model or epoch selection.
Its gold-interval test accuracy is 36.72\%. The primary validation selectors
remain frozen, and the second consumer evaluates their held-out outputs.

\subsection{Evaluation and reproducibility}
We report localization, coverage, retained evidence, temporal purity,
prediction/reference count, matched and all-reference accuracy, AUC-RU,
and substitution/deletion/insertion components for \WERall and \WERiv.
The rejection sensitivity uses the same frozen RGB logits and does not retrain
either segmenter or recognizer. Validation fixes every selector and confidence
threshold before test evaluation. The supplementary material provides
implementation details, full per-seed results, protocol specifications, and
robustness analyses. We will release split and vocabulary manifests,
perturbation-generation code, candidate predictions, frozen-selection manifests,
per-video results, and table-generation scripts. Corpus videos remain with their
providers.

\section{Results}
\label{sec:results-v2}

\subsection{Equal overlap permits different recognition outcomes}
Overlap predicts recognition well on average. Across the DGS controlled-edit
matrix, Spearman $\rho$ between \mFoneS and consumer accuracy is $.815$ for the
statistical pose consumer and $.808$ for the sequence consumer. Averages hide
the cases that decide model selection. Compared pairwise, \mFoneS and
recognition disagree on roughly 19\% of DGS pairs, and on 24.0\% and 23.9\% of
Dicta-Sign pairs for the two consumers. Requiring at least one accuracy point
of separation removes near-ties and leaves 14.5\% on DGS and 19.2\%/19.4\% on
Dicta-Sign. Roughly one comparison in six therefore remains a substantive
disagreement.

\begin{figure*}[t]
  \centering
  \includegraphics[width=.99\textwidth,trim=0 35 0 0,clip]{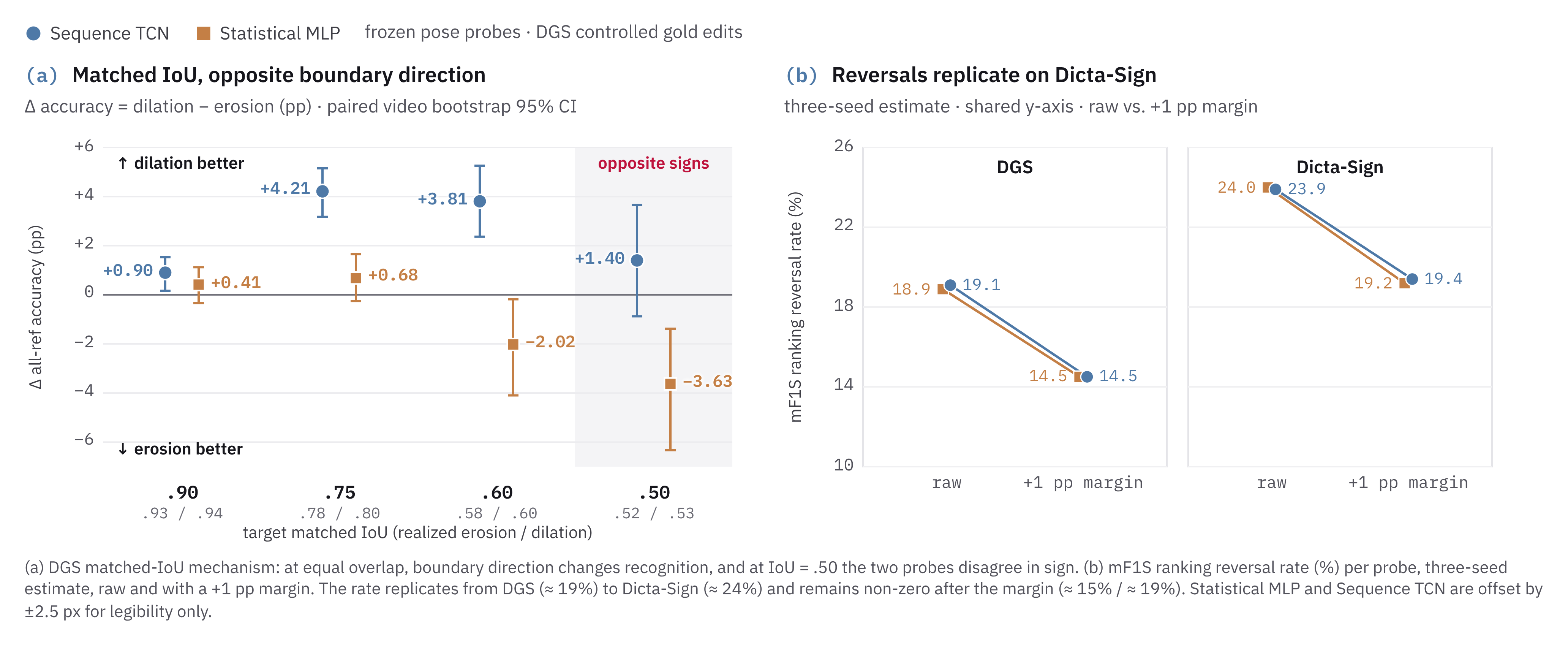}
  \caption{Controlled mechanism and replication. (a) Dilation-minus-erosion
  recognition difference at matched IoU; whiskers are paired video-bootstrap
  95\% intervals. (b) Mean per-seed \mFoneS ranking-reversal rates before and
  after a one-point practical recognition margin.}
  \label{fig:controlled}
\end{figure*}

Figure~\ref{fig:controlled}a isolates the boundary mechanism. At target IoU
$.5$, the inner interval retains half of the reference, whereas the outer
interval contains the reference plus an equal-IoU amount of context. The
statistical consumer favors erosion by 3.6 points; the sequence consumer favors
dilation by 1.4 points. A strict exact-IoU subset preserves both directions
($-4.44$ and $+2.40$ points). Equal-frame pairs reveal the role of edit
magnitude: after removing or adding four frames, both consumers favor dilation
by 3.82 and 10.79 points. IoU therefore combines two quantities with distinct
consumer costs, retained evidence $C$ and purity $Q$. Equal IoU is not equal
input to a recognizer.

The controlled edits explain the mechanism. We next ask whether it changes the
choice among real segmenters.

\subsection{Overlap and reference recognition align except on BSL}
\begin{table*}[t]
\centering
\caption{Held-out results for all 27 real-output conditions. Each cell lists
seeds $7/42/2027$; bold marks within-family validation selection for \mFoneS,
AUC-RU, and the primary \WERall objective. Coverage and accuracy preserve every
seed for inspection.}
\label{tab:seed-matrix}
\footnotesize
\setlength{\tabcolsep}{2.4pt}
\begin{tabular}{llccccc}
\toprule
\textbf{Corpus} & \textbf{Family} & \textbf{\mFoneS} & \textbf{Coverage} & \textbf{Accuracy} & \textbf{AUC-RU} & \textbf{\WERall} \\
\midrule
DGS & TCN & 36.5 / \textbf{38.3} / 37.9 & 95.9 / 96.0 / 96.3 & 37.0 / 38.2 / 37.7 & 0.259 / 0.273 / \textbf{0.268} & 272.18 / 272.73 / \textbf{271.36} \\
 & BiLSTM & 43.8 / \textbf{44.2} / 42.7 & 95.8 / 96.0 / 96.4 & 39.5 / 39.8 / 39.8 & 0.306 / \textbf{0.305} / 0.297 & 248.47 / \textbf{241.23} / 246.53 \\
 & Transformer & 39.5 / 39.9 / \textbf{40.2} & 95.7 / 96.3 / 95.6 & 39.0 / 39.7 / 40.0 & 0.285 / \textbf{0.297} / 0.294 & 274.58 / 274.45 / \textbf{269.45} \\
\midrule
Dicta-Sign & TCN & \textbf{32.7} / 28.6 / 30.3 & 90.3 / 84.1 / 89.2 & 42.1 / 30.6 / 39.3 & \textbf{0.259} / 0.152 / 0.227 & 302.28 / \textbf{253.22} / 287.96 \\
 & BiLSTM & \textbf{35.6} / 32.4 / 30.8 & 87.9 / 92.9 / 89.4 & 45.1 / 42.8 / 38.4 & \textbf{0.298} / 0.251 / 0.211 & 305.85 / 335.93 / \textbf{290.63} \\
 & Transformer & 21.3 / \textbf{22.3} / 20.9 & 82.3 / 87.8 / 80.2 & 30.1 / 35.9 / 28.4 & 0.135 / \textbf{0.175} / 0.119 & 273.49 / 348.61 / \textbf{240.54} \\
\midrule
BSL & TCN & 56.5 / \textbf{57.6} / 58.6 & 80.6 / 80.3 / 78.3 & 24.1 / 24.4 / 23.9 & 0.130 / \textbf{0.132} / 0.128 & 97.91 / 97.91 / \textbf{93.74} \\
 & BiLSTM & 58.0 / 57.1 / \textbf{60.4} & 78.1 / 77.1 / 74.7 & 23.7 / 23.4 / 22.4 & 0.123 / 0.122 / \textbf{0.121} & 90.45 / 90.18 / \textbf{88.60} \\
 & Transformer & \textbf{43.0} / 45.1 / 43.2 & 83.1 / 82.5 / 83.3 & 24.7 / 25.0 / 25.2 & \textbf{0.121} / 0.126 / 0.132 & 117.71 / 116.39 / \textbf{116.77} \\
\bottomrule
\end{tabular}
\end{table*}

\begin{figure*}[!t]
  \centering
  \includegraphics[width=.99\textwidth,trim=0 29 0 0,clip]{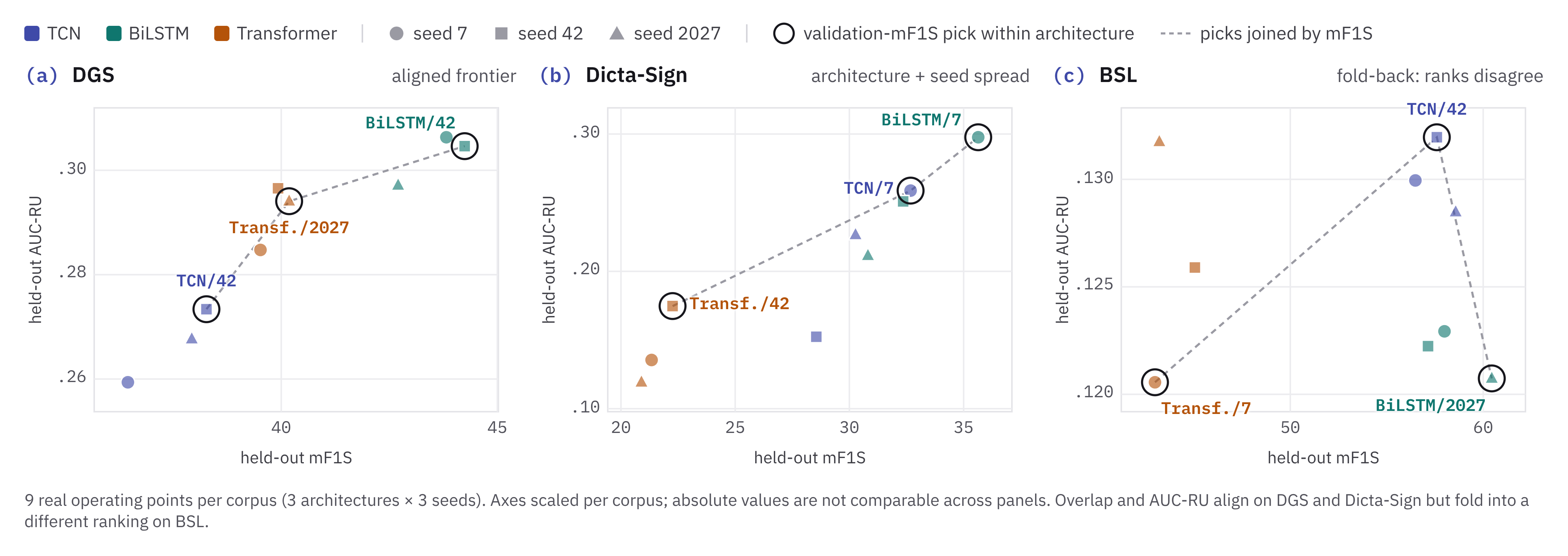}
  \caption{Held-out localization and reference utility for all 27 real-output
  conditions. Color denotes architecture and marker denotes seed. Black rings
  identify the validation-\mFoneS seed within each family. DGS and Dicta-Sign
  form aligned frontiers; BSL separates overlap from reference utility.}
  \label{fig:architecture-map}
\end{figure*}

Figure~\ref{fig:architecture-map} compares localization and recognition, while
Table~\ref{tab:seed-matrix} reports all three seeds for every architecture.
DGS forms compact family clusters. Validation-\mFoneS ranks BiLSTM above
Transformer above TCN, and AUC-RU gives the same family order. Dicta-Sign also
places BiLSTM first under both criteria. Together they show the common regime:
localization and recognition improve together.

BSL separates the axes. Its validation-selected BiLSTM reaches the highest
\mFoneS (60.45) and lowest \WERiv (88.12), but covers 74.71\% of references and
recognizes 22.44\%. The Transformer reaches 83.08\% coverage and 24.72\%
accuracy at 42.99 \mFoneS. TCN lies between them in localization and achieves
the marginally highest AUC-RU (.13196 versus .13175 for Transformer/2027). No
single BSL model maximizes overlap, reference
recovery, and overlap-conditioned recognition.

Candidate pairs tell the same story. With a
one-point accuracy margin, \mFoneS reverses accuracy in 0/19 DGS, 1/33
Dicta-Sign, and 16/17 BSL comparisons. The BSL conflict therefore survives the
practical margin, while small DGS differences disappear. Table~\ref{tab:selection-costs}
turns these disagreements into held-out model-selection costs. All three
objectives select the same DGS model. Dicta-Sign and BSL each contain one
objective conflict, in opposite directions.

\begin{table*}[t]
\centering
\caption{Held-out change relative to the validation-\mFoneS selection. Every model is selected on validation; positive coverage, accuracy, and AUC-RU are benefits, while negative \WERall is a benefit.}
\label{tab:selection-costs}
\footnotesize
\setlength{\tabcolsep}{5.0pt}
\begin{tabular}{lllrrrr}
\toprule
\textbf{Corpus} & \textbf{Validation objective} & \textbf{Selected} & \textbf{$\Delta$Cov.} & \textbf{$\Delta$Acc.} & \textbf{$\Delta$AUC-RU} & \textbf{$\Delta$\WERall} \\
\midrule
DGS & AUC-RU & BiLSTM/42 & +0.00 & +0.00 & +0.0000 & +0.00 \\
 & \WERall & BiLSTM/42 & +0.00 & +0.00 & +0.0000 & +0.00 \\
\addlinespace[1.5pt]
Dicta-Sign & AUC-RU & BiLSTM/7 & +0.00 & +0.00 & +0.0000 & +0.00 \\
 & \WERall & TCN/42 & -3.77 & -14.52 & -0.1453 & -52.63 \\
\addlinespace[1.5pt]
BSL & AUC-RU & TCN/42 & +5.62 & +1.97 & +0.0112 & +9.31 \\
 & \WERall & BiLSTM/2027 & +0.00 & +0.00 & +0.0000 & +0.00 \\
\bottomrule
\end{tabular}
\end{table*}

\subsection{Recognizer changes preserve the main conclusions}
The DGS outputs were evaluated by three independently trained recognizers.
Table~\ref{tab:dgs-consumers} shows a stable architecture-level result. Rank
correlations between \mFoneS and AUC-RU range from $.933$ to $.983$, and
every recognizer selects a BiLSTM. The statistical MLP selects seed 7 instead
of seed 42; on test, that seed change gives $-0.58$ accuracy points
$[-2.31,+0.85]$ and $-.0017$ AUC-RU $[-.0129,+.0076]$. All three recognizers
therefore agree on the top-ranked architecture for DGS.

\begin{table}[t]
\centering
\caption{The same nine DGS outputs evaluated by three frozen consumers. $\rho$
compares \mFoneS and AUC-RU; Rev. applies a one-point accuracy margin.}
\label{tab:dgs-consumers}
\scriptsize
\setlength{\tabcolsep}{2.4pt}
\begin{tabular}{lcccc}
\toprule
\textbf{Consumer} & \textbf{Gold acc.} & \textbf{$\rho$} & \textbf{Rev.} & \textbf{Val. AUC-RU sel.} \\
\midrule
RGB Logos & 46.88 & .967 & 0/19 & BiLSTM/42 \\
Statistical MLP & 35.75 & .983 & 0/15 & BiLSTM/7 \\
Sequence TCN & 39.84 & .933 & 1/20 & BiLSTM/42 \\
\bottomrule
\end{tabular}
\end{table}

\begin{figure}[t]
  \centering
  \includegraphics[width=\columnwidth]{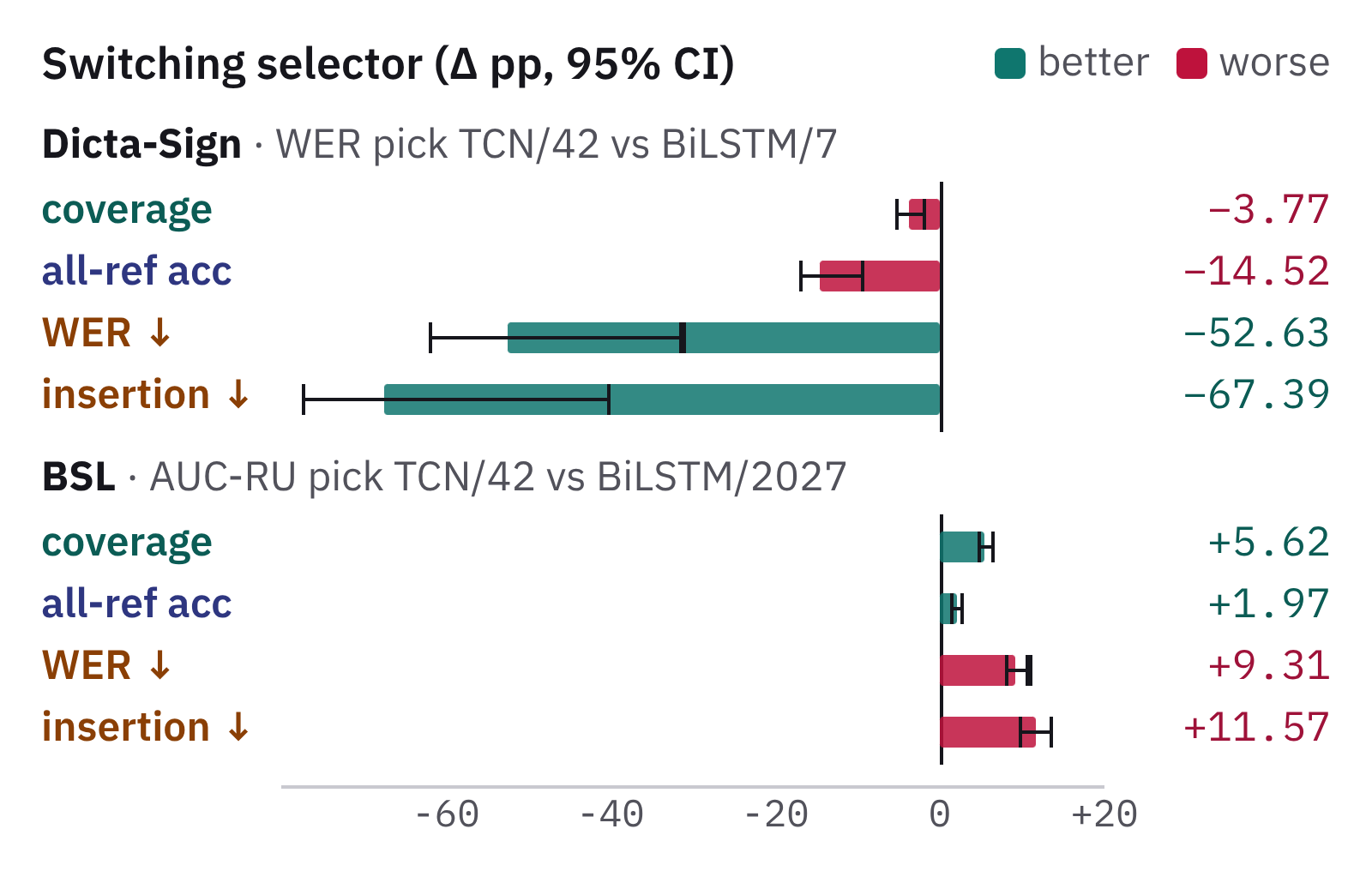}
  \caption{Held-out changes from switching the validation objective on
  Dicta-Sign and BSL. Whiskers are 95\% signer-clustered intervals; green and
  red indicate benefit and cost after orienting each metric.}
  \label{fig:selector-tradeoffs}
\end{figure}

BSL provides a clear objective trade-off. Validation \mFoneS and \WERall
select BiLSTM/2027; AUC-RU selects TCN/42. Relative to BiLSTM, TCN gains 5.62
coverage points, 1.97 accuracy points, and .0112 AUC-RU. The same choice adds
9.31 \WERall points, including 11.57 more insertion points. Every
signer-clustered interval excludes zero (Fig.~\ref{fig:selector-tradeoffs}).
Reference recovery and complete-sequence error favor different systems.

The independent BSL-1K I3D--TCN consumer reproduces the preference
(Table~\ref{tab:bsl-consumers}). TCN/42 gains 2.14 all-reference accuracy
points $[1.45,2.81]$ and .0114 AUC-RU $[.0058,.0169]$ over BiLSTM/2027.
All 20 candidate pairs separated by at least one accuracy point
reverse the \mFoneS ordering. The BSL conflict therefore persists across
VideoSwin and I3D visual representations and across MLP and TCN recognition
heads.

\begin{table}[t]
\centering
\caption{BSL conflict replication across independently trained RGB consumers.
Rev. counts \mFoneS--accuracy ranking reversals among candidate pairs separated
by at least one accuracy point. $\Delta$ compares the validation-frozen TCN/42
and BiLSTM/2027 outputs.}
\label{tab:bsl-consumers}
\footnotesize
\setlength{\tabcolsep}{3.2pt}
\begin{tabular}{lcccc}
\toprule
\textbf{Consumer} & \textbf{Gold Top-1} & \textbf{Rev.} & \textbf{$\Delta$Acc.} & \textbf{$\Delta$AUC-RU} \\
\midrule
VideoSwin & 40.82 & 16/17 & $+1.97$ & $+.0112$ \\
BSL-1K I3D--TCN & 36.72 & 20/20 & $+2.14$ & $+.0114$ \\
\bottomrule
\end{tabular}
\end{table}

\subsection{Sequence error exposes output burden}
Table~\ref{tab:wer-scope} separates all-output and in-vocabulary sequence error.
On DGS and Dicta-Sign, 31--42\% of predicted clips align more strongly with an
annotated OOV sign than with any in-vocabulary reference. Removing those clips
reduces absolute WER substantially. The residual \WERiv remains above 100\%, so
insertion and recognition errors remain within the evaluable vocabulary. BSL
changes by less than one point because its vocabulary covers
nearly every annotated lexical label.

\begin{table}[t]
\centering
\caption{Vocabulary and sequence burden on held-out test. OOV pred. is the
fraction assigned to annotated OOV signs. The primary \WERall selector and the
\WERiv sensitivity analysis select the same candidate in every corpus.}
\label{tab:wer-scope}
\scriptsize
\setlength{\tabcolsep}{2.5pt}
\begin{tabular}{lcccc}
\toprule
\textbf{Corpus} & \textbf{OOV pred.} & \textbf{\WERall} & \textbf{\WERiv} & \textbf{Selector} \\
\midrule
DGS & 31--33\% & 241--275 & 150--175 & BiLSTM/42 \\
Dicta-Sign & 39--42\% & 241--349 & 134--201 & TCN/42 \\
BSL & .4--.7\% & 89--118 & 88--117 & BiLSTM/2027 \\
\bottomrule
\end{tabular}
\end{table}

Both sequence protocols select the same candidate on Dicta-Sign. \WERall and
\WERiv select TCN/42, whereas \mFoneS and AUC-RU select BiLSTM/7.
TCN lowers held-out \WERall by 52.63 points and \WERiv by 31.17 points. The
price is 3.77 coverage points, 14.52 accuracy points, and
.1453 AUC-RU. Every signer-clustered interval excludes zero
(Fig.~\ref{fig:selector-tradeoffs}), and a 19-recording bootstrap preserves the
\WERiv reduction $[-41.98,-18.20]$. Lower sequence error therefore does not
imply that more reference signs survive the pipeline.

Removing the low-overlap Transformer family does not change the Dicta-Sign
decision. Within the six TCN/BiLSTM candidates, validation \mFoneS still selects
BiLSTM/7, validation \WERall still selects TCN/42, and 13/15 one-point WER
comparisons reverse the overlap ordering. The conflict arises within comparable
model families.

The AUC-RU selector is also stable to its overlap grid. Integrating $.3$--$.7$,
integrating $.4$--$.6$, or using $U(.5)$ selects BiLSTM/42 on DGS, BiLSTM/7 on
Dicta-Sign, and TCN/42 on BSL. Direct all-reference accuracy selects a different
candidate on DGS and BSL. AUC-RU adds overlap strictness to recognition
correctness; it is not a renamed accuracy score.

Confidence rejection does not remove the conflict.
\mbox{Table~\ref{tab:rejection}} fixes one threshold per corpus to retain 90\% of
the shared validation prediction cache. Rejection lowers held-out \WERiv, but the selected
models remain BiLSTM/42 on DGS, TCN/42 on Dicta-Sign, and BiLSTM/2027 on BSL.
The same models are selected at 95\% and 80\% retention.

\begin{table}[t]
\centering
\caption{Validation-frozen rejection sensitivity. The 90\% operating point sets one corpus-level confidence threshold from validation. Values are held-out \WERiv/coverage (\%).}
\label{tab:rejection}
\footnotesize
\setlength{\tabcolsep}{3.2pt}
\begin{tabular}{lccc}
\toprule
\textbf{Corpus} & \textbf{No rejection} & \textbf{90\% retention} & \textbf{Selected} \\
\midrule
DGS & 150.26/96.0 & 130.68/89.2 & BiLSTM/42 \\
Dicta-Sign & 137.96/84.1 & 124.73/77.4 & TCN/42 \\
BSL & 88.12/74.7 & 84.74/70.1 & BiLSTM/2027 \\
\bottomrule
\end{tabular}
\end{table}

Without a retention constraint, WER rewards silence. Jointly selecting the
candidate and threshold by validation \WERiv retains only 25.9\%, 26.8\%, and
66.1\% of the original DGS, Dicta-Sign, and BSL predictions. Dicta-Sign then
switches to BiLSTM/7, but coverage falls to 40.3\%. Rejection controls
insertions; coverage prevents it from discarding useful output.

\section{Discussion}

\subsection{What should a segmentation benchmark report?}
\mFoneB and \mFoneS tell us where a predicted segment lies relative to a manual
annotation. They do not tell us whether the clip still contains enough visual
evidence for recognition or whether the complete output sequence contains too
many errors. Our experiments separate these questions into three measurements:
reference recovery, clip recognition, and sequence error.

AUC-RU asks how many annotated signs are both recovered and recognized. WER
asks whether the complete predicted sequence is correct and counts extra clips
as insertions. These measures answer different practical questions. Retrieval
and annotation tools may value recovering more signs, whereas transcription
systems may prefer fewer sequence errors. A model can improve one goal while
hurting the other.

The model-selection results make this distinction concrete. On BSL, \mFoneS
selects the same model as \WERall; on Dicta-Sign, it selects the same model as
AUC-RU. The downstream objectives disagree, so no single overlap ordering
represents both across corpora.

A benchmark should therefore state which downstream system will use the
segments and freeze that system before evaluation. It should report four sets
of results: localization; reference coverage and retained content; recognition;
and sequence error with substitution, deletion, and insertion counts. The
validation metric should match the intended application. The other results show
what that choice sacrifices.

\subsection{Recognition tests the content inside a boundary}
Manual intervals remain the reference for temporal localization. A frozen
recognizer tests whether the frames inside each interval still show the sign.
Two intervals can receive the same IoU while retaining different visual
evidence. Annotators can also place a co-articulated transition differently
while preserving the same recognizable sign. All-reference accuracy and AUC-RU
count annotated signs that remain recognizable.

We also test whether the choice of recognizer changes the conclusion on DGS.
Absolute accuracy changes across RGB, pooled-pose, and temporal-pose models,
but every independently trained RGB and pose model ranks BiLSTM first. At
finer resolution, the statistical MLP
changes the selected BiLSTM seed, but the held-out interval crosses zero. The
top-ranked architecture is stable; seed-level claims remain tied to the
declared consumer.

BSL shows the complementary case. Two independently trained RGB consumers use
different visual backbones and temporal heads, yet both prefer TCN/42 to
BiLSTM/2027 for reference recognition. Agreement across the two models shows
that the BSL coverage gain is not unique to one recognizer.

\subsection{Limited vocabularies explain part of the high WER}
The train-supported consumer vocabularies cover only 53.9\% of all annotated
DGS test tokens and 43.4\% of all annotated Dicta-Sign test tokens, while the
segmenters emit clips throughout each video. \WERall scores the complete
prediction stream against the in-vocabulary
reference sequence. \WERiv uses the same reference sequence but removes
predictions whose strongest overlap is with an annotated OOV sign. The
difference between the two measures therefore reflects vocabulary coverage.
Errors shared by both measures come from missed signs, fragmented signs, false
predictions, and recognition substitutions.

WER is easiest to interpret on BSL, where \WERiv spans 88--117\%. On DGS and
Dicta-Sign it remains above 100\%, so it is most useful for comparing models
under the same fixed recognizer. Coverage, recognition, and the separate
substitution, deletion, and insertion counts explain each difference.

The count ratios explain the high WER values. DGS models
emit 1.10--1.23 times as many segments as the full gold annotation, yet the
closed-vocabulary hypothesis contains 2.85--3.19 times as many clips as the
in-vocabulary reference. The first ratio reflects segmentation behavior; the
second combines segmentation with consumer coverage. Together, the two ratios
separate segmentation behavior from consumer-vocabulary coverage.

Confidence rejection removes some insertion errors without changing the
segmenter output. Moderate rejection lowers WER without changing any
WER-selected model. Selecting the candidate and rejection threshold by
validation WER alone discards most DGS and Dicta-Sign predictions. Sequence
error must therefore be interpreted together with coverage.

\subsection{Relation to structured segmentation}
BEACON combines begin/end-aware supervision, boundary distillation, and
structured decoding for continuous sign segmentation \cite{zhang2027beacon}.
Its contribution is a segmenter that produces well-formed outputs. The
complementary evaluation question is which measurements identify useful clips
after a segmenter emits them. The common argmax pool isolates three
architecture families under one readout. Structured decoders can enter the
same benchmark as additional participants. The same measures will show whether
reduced fragmentation improves recognition, sequence error, or both.

\subsection{Limitations and scope}
This protocol targets pipelines that produce explicit lexical clips:
recognition-by-clipping, retrieval, annotation, and alignment. Continuous
recognition and translation consume the full video stream and require a
different interface. The
consumer vocabularies emphasize frequent lexical signs represented in
training. Open-vocabulary and category-aware consumers extend the protocol to
rare signs, fingerspelling, pointing, and depicting constructions.

The downstream experiments use isolated-sign recognizers. They directly test
clip recognition and segment-then-recognize WER, but not retrieval quality,
subtitle alignment, or human annotation time. Those applications require their
own frozen consumers and task-specific outcomes. The three corpora also differ
in vocabulary coverage and annotation protocol, so comparisons remain within
corpus.

The benchmark spans three architectures, three seeds, three RGB
corpus protocols, two additional DGS consumers, and controlled pose consumers.
Structured decoders, pretrained video segmenters, and retrieval or human-effort
consumers can enter the same protocol. The rule is simple: freeze the downstream
system, evaluate identical segment outputs, and report localization, recovery,
recognition, and sequence error.

\section{Conclusion}
Sign segmentation determines both where a clip lies and what a downstream
system receives. Across three corpora, 27 real model outputs, multiple
frozen recognizers, and controlled boundary edits, overlap predicts recognition
on average but does not always select the same model. DGS shows agreement across
recognizers. On BSL, \mFoneS selects the WER-preferred model; on Dicta-Sign, it
selects the AUC-RU-preferred model. The downstream objectives disagree in both
corpora. Vocabulary-aware scoring reduces the absolute WER on DGS and
Dicta-Sign without changing the selected model. A benchmark should
therefore report temporal overlap together with reference recovery,
recognition, and sequence error.

\section{Ethical Impact Statement}
This study performs secondary analysis of the Public DGS Corpus,
Dicta-Sign-LSF-v2, the BSL Corpus, and BOBSL-derived recognition resources
under their research-use conditions. The study collects no new participant
data and redistributes neither source videos nor reusable identity-bearing
feature archives. Aggregate reporting follows each corpus's documented
annotation scope and conventions.

\FloatBarrier
{\footnotesize
\bibliographystyle{ieee}
\bibliography{references}

\begin{thebibliography}{10}\itemsep=-1pt

\bibitem{farha2019mstcn}
Y.~Abu~Farha and J.~Gall.
\newblock {MS-TCN}: Multi-stage temporal convolutional network for action
  segmentation.
\newblock In {\em IEEE Conference on Computer Vision and Pattern Recognition},
  pages 3575--3584, 2019.

\bibitem{albanie2021bobsl}
S.~Albanie, G.~Varol, L.~Momeni, H.~Bull, T.~Afouras, H.~Chowdhury, N.~Fox,
  B.~Woll, R.~Cooper, A.~McParland, and A.~Zisserman.
\newblock {BBC-Oxford British Sign Language Dataset}.
\newblock {\em arXiv preprint arXiv:2111.03635}, 2021.

\bibitem{belissen2020dictasign}
V.~Belissen, A.~Braffort, and M.~Gouiff{\`e}s.
\newblock Dicta-sign-{LSF}-v2: Remake of a continuous {F}rench {S}ign
  {L}anguage dialogue corpus and a first baseline for automatic sign language
  processing.
\newblock In {\em Proceedings of the 12th Language Resources and Evaluation
  Conference (LREC)}, pages 6040--6048, Marseille, France, 2020. European
  Language Resources Association (ELRA).

\bibitem{bull2020subtitle}
H.~Bull, M.~Gouiff{`e}s, and A.~Braffort.
\newblock Automatic segmentation of sign language into subtitle-units.
\newblock In {\em Computer Vision -- ECCV 2020 Workshops}, volume 12536 of {\em
  Lecture Notes in Computer Science}, pages 186--198. Springer, 2020.

\bibitem{chavelli2025interpretable}
F.~Chavelli, P.~Boniol, and M.~Thomazo.
\newblock Toward interpretable evaluation measures for time series
  segmentation.
\newblock In {\em Advances in Neural Information Processing Systems},
  volume~38, 2025.

\bibitem{chen2022twostream}
Y.~Chen, R.~Zuo, F.~Wei, Y.~Wu, S.~Liu, and B.~Mak.
\newblock Two-stream network for sign language recognition and translation.
\newblock In {\em Advances in Neural Information Processing Systems},
  volume~35, 2022.

\bibitem{gueuwou2025shubert}
S.~Gueuwou, X.~Du, G.~Shakhnarovich, K.~Livescu, and A.~H. Liu.
\newblock {SHuBERT}: Self-supervised sign language representation learning via
  multi-stream cluster prediction.
\newblock In {\em Proceedings of the 63rd Annual Meeting of the Association for
  Computational Linguistics}, pages 28792--28810, 2025.

\bibitem{hanke2020dgscorpus}
T.~Hanke, M.~Schulder, R.~Konrad, and E.~Jahn.
\newblock Extending the {P}ublic {DGS} {C}orpus in size and depth.
\newblock In {\em Proceedings of the LREC2020 9th Workshop on the
  Representation and Processing of Sign Languages}, pages 75--82, Marseille,
  France, 2020. European Language Resources Association (ELRA).

\bibitem{ishikawa2021boundary}
Y.~Ishikawa, S.~Kasai, Y.~Aoki, and H.~Kataoka.
\newblock Alleviating over-segmentation errors by detecting action boundaries.
\newblock In {\em IEEE/CVF Winter Conference on Applications of Computer
  Vision}, pages 2322--2331, 2021.

\bibitem{jiang2026sea}
Z.~Jiang, Y.~Jang, L.~Momeni, G.~Varol, S.~Ebling, and A.~Zisserman.
\newblock Segment, embed, and align: A universal recipe for aligning subtitles
  to signing.
\newblock In {\em Proceedings of the 64th Annual Meeting of the Association for
  Computational Linguistics}, pages 30371--30384, 2026.

\bibitem{jiang2025meaningful}
Z.~Jiang, C.~Leong, A.~Moryossef, O.~Cory, M.~Ivashechkin, N.~Tarigopula,
  B.~Zhang, A.~G{\"o}hring, A.~Rios, R.~Sennrich, and S.~Ebling.
\newblock Meaningful pose-based sign language evaluation.
\newblock In {\em Proceedings of the Tenth Conference on Machine Translation
  (WMT)}, pages 64--80, 2025.

\bibitem{lea2017tcn}
C.~Lea, M.~D. Flynn, R.~Vidal, A.~Reiter, and G.~D. Hager.
\newblock Temporal convolutional networks for action segmentation and
  detection.
\newblock In {\em IEEE Conference on Computer Vision and Pattern Recognition},
  pages 156--165, 2017.

\bibitem{li2024otas}
Y.~Li, Z.~Xue, and H.~Xu.
\newblock {OTAS}: Unsupervised boundary detection for object-centric temporal
  action segmentation.
\newblock In {\em IEEE/CVF Winter Conference on Applications of Computer
  Vision}, pages 6437--6446, 2024.

\bibitem{li2021timestamp}
Z.~Li, Y.~Abu~Farha, and J.~Gall.
\newblock Temporal action segmentation from timestamp supervision.
\newblock In {\em IEEE/CVF Conference on Computer Vision and Pattern
  Recognition}, pages 8365--8374, 2021.

\bibitem{low2025handson}
J.~Low, H.~Walsh, O.~M. Sincan, and R.~Bowden.
\newblock {Hands-On}: Segmenting individual signs from continuous sequences.
\newblock In {\em IEEE International Conference on Automatic Face and Gesture
  Recognition (FG)}, 2025.

\bibitem{moryossef2023seg}
A.~Moryossef, Z.~Jiang, M.~M{\"u}ller, S.~Ebling, and Y.~Goldberg.
\newblock Linguistically motivated sign language segmentation.
\newblock In {\em Findings of the Association for Computational Linguistics:
  EMNLP 2023}, pages 12703--12724, 2023.

\bibitem{ovodov2025logos}
I.~Ovodov, P.~Surovtsev, K.~Kvanchiani, A.~Kapitanov, and A.~Nagaev.
\newblock Logos as a well-tempered pre-train for sign language recognition.
\newblock In {\em Proceedings of the 2025 Conference on Empirical Methods in
  Natural Language Processing}, pages 24340--24353, 2025.

\bibitem{renz2021segmentation}
K.~Renz, N.~C. Stache, S.~Albanie, and G.~Varol.
\newblock Sign language segmentation with temporal convolutional networks.
\newblock In {\em IEEE International Conference on Acoustics, Speech and Signal
  Processing (ICASSP)}, pages 2135--2139, 2021.

\bibitem{renz2021changepoint}
K.~Renz, N.~C. Stache, N.~Fox, G.~Varol, and S.~Albanie.
\newblock Sign segmentation with changepoint-modulated pseudo-labelling.
\newblock In {\em IEEE/CVF Conference on Computer Vision and Pattern
  Recognition Workshops}, pages 3403--3412, 2021.

\bibitem{schembri2013bsl}
A.~Schembri, J.~Fenlon, R.~Rentelis, S.~Reynolds, and K.~Cormier.
\newblock Building the {British Sign Language Corpus}.
\newblock {\em Language Documentation and Conservation}, 7:136--154, 2013.

\bibitem{varol2021read}
G.~Varol, L.~Momeni, S.~Albanie, T.~Afouras, and A.~Zisserman.
\newblock Read and attend: Temporal localisation in sign language videos.
\newblock In {\em IEEE/CVF Conference on Computer Vision and Pattern
  Recognition}, pages 16857--16866, 2021.

\bibitem{woll2022humans}
B.~Woll, N.~Fox, and K.~Cormier.
\newblock Segmentation of signs for research purposes: Comparing humans and
  machines.
\newblock In {\em Proceedings of the 10th Workshop on the Representation and
  Processing of Sign Languages}, pages 198--201, 2022.

\bibitem{zhang2023sltunet}
B.~Zhang, M.~M{\"u}ller, and R.~Sennrich.
\newblock {SLTUNET}: A simple unified model for sign language translation.
\newblock In {\em The Eleventh International Conference on Learning
  Representations (ICLR)}, Kigali, Rwanda, 2023.

\bibitem{zhang2027beacon}
G.~Zhang and K.~Stefanov.
\newblock {BEACON}: Distilled begin/end-aware structured decoding for
  continuous sign language segmentation.
\newblock {\em arXiv preprint arXiv:2610.39553}, 2026.
\newblock Accepted at the IEEE/CVF Winter Conference on Applications of
  Computer Vision (WACV) 2027.

\bibitem{zhao2026mhb}
M.~Zhao, Z.~Yang, Y.~Zhou, Z.~Xia, C.~Jin, X.~He, S.~Lin, C.~Neidle, and
  D.~Metaxas.
\newblock Continuous sign language recognition using multimodal input and
  handshape-aware boundary detection.
\newblock In {\em Proceedings of the {LREC} 2026 12th Workshop on the
  Representation and Processing of Sign Languages: Language in Motion}, pages
  501--512, 2026.

\bibitem{zhou2023gfslt}
B.~Zhou, Z.~Chen, A.~Clap{\'e}s, J.~Wan, Y.~Liang, S.~Escalera, Z.~Lei, and
  D.~Zhang.
\newblock Gloss-free sign language translation: Improving from visual--language
  pretraining.
\newblock In {\em IEEE/CVF International Conference on Computer Vision}, pages
  20871--20881, 2023.

\bibitem{zuo2024online}
R.~Zuo, F.~Wei, and B.~Mak.
\newblock Towards online continuous sign language recognition and translation.
\newblock In {\em Proceedings of the 2024 Conference on Empirical Methods in
  Natural Language Processing}, pages 11050--11067, 2024.

\end{thebibliography}
}

\end{document}